%% file: main.tex
\documentclass[10pt,twocolumn,letterpaper]{article}

\usepackage[pagenumbers]{cvpr} 

\input{preamble}

\definecolor{cvprblue}{rgb}{0.21,0.49,0.74}
\usepackage[pagebackref,breaklinks,colorlinks,citecolor=cvprblue]{hyperref}

\def\paperID{*****} 
\def\confName{CVPR}
\def\confYear{2024}

\title{GPT4Vis: What Can GPT-4 Do for Zero-shot Visual Recognition?}

\author{%
Wenhao Wu$^{1,2}\textsuperscript{* \Envelope}$,
Huanjin Yao$^{2,3}\textsuperscript{*}$,
Mengxi Zhang$^{2,4}\textsuperscript{*}$,
Yuxin Song$^{2}$,
Wanli Ouyang$^{5}$,
Jingdong Wang$^{2}$\\
$^1$The University of Sydney \quad $^2$Baidu Inc.  \quad
$^3$Tsinghua University  \\ 
$^4$Tianjin University \quad $^5$The Chinese University of Hong Kong \\
{\tt\small \url{https://github.com/whwu95/GPT4Vis}}
}

\begin{document}
\maketitle

\epigraph{\textit{``The world as we have created it is a process of our thinking. It cannot be changed without changing our thinking.''}}{--- \textit{Albert Einstein}}

\input{sec/0_abstract}    
\input{sec/1_intro}

\input{sec/5_related}

\input{sec/2_apporach}

\input{sec/3_experiment}

\input{sec/4_limitation}

{
    \small
    \bibliographystyle{ieeenat_fullname}
    \bibliography{main}
}


\end{document}

%% file: preamble.tex
\usepackage[dvipsnames]{xcolor}
\usepackage{colortbl}

\usepackage{epigraph}
\usepackage{bbding}
\usepackage{nicematrix}     
\definecolor{00red}{RGB}{236,35,35}
\newcommand{\whred}[1]{\textcolor{00red}{#1}}
\definecolor{00blue}{RGB}{50,149,237}
\newcommand{\whblue}[1]{\textcolor{00blue!80}{#1}}
\definecolor{00green}{RGB}{82,208,83}

\definecolor{02pink}{RGB}{240,178,188}
\newcommand{\rpink}{\rowcolor{02pink!40}}
\newcommand{\gpt}[1]{\cellcolor{00blue!15}{#1}}
\definecolor{demphcolor1}{gray}{.6}

\usepackage{indentfirst}
\usepackage{multirow}
\usepackage{fontawesome}
\newcommand*\cbottomrule[1]{\cmidrule[\heavyrulewidth]{#1}}
\newcommand*\ctoprule[1]{\addlinespace[0.7mm]\cmidrule[\heavyrulewidth]{#1}}
\newcommand\blfootnote[1]{%
  \begingroup
  \renewcommand\thefootnote{}\footnote{#1}%
  \addtocounter{footnote}{-1}%
  \endgroup
}
\def\x{$\times$}

%% file: sec/0_abstract.tex
\begin{abstract}
This paper does not present a novel method. Instead, it delves into an essential, yet must-know baseline in light of the latest advancements in Generative Artificial Intelligence (GenAI): the utilization of GPT-4 for visual understanding. 
Our study centers on the evaluation of GPT-4's \textbf{\whblue{linguistic}} and \textbf{\whred{visual}} capabilities in zero-shot visual recognition tasks: Firstly, we explore the potential of its generated rich textual descriptions across various categories to enhance recognition performance without any training. Secondly, we evaluate GPT-4's visual proficiency in directly recognizing diverse visual content. 
We conducted extensive experiments to systematically evaluate GPT-4's performance across images, videos, and point clouds, using 16 benchmark datasets to measure top-1 and top-5 accuracy. 
Our findings show that GPT-4, enhanced with rich linguistic descriptions, significantly improves zero-shot recognition, offering an average top-1 accuracy increase of 7\% across all datasets. 
GPT-4 excels in visual recognition, outshining OpenAI-CLIP's ViT-L and rivaling EVA-CLIP's ViT-E, particularly in video datasets HMDB-51 and UCF-101, where it leads by 22\% and 9\%, respectively.
We hope this research contributes valuable data points and experience for future studies.
We release our code at \url{https://github.com/whwu95/GPT4Vis}.
\end{abstract}

%% file: sec/1_intro.tex
\section{Introduction}
\label{sec:intro}
\blfootnote{*~Research interns at Baidu \quad  \Envelope~Corresponding author} 
ChatGPT~\cite{chatgpt}, lanuched in November 2022, marked a seismic shift in the application of AI, sparking a ``wow'' moment that galvanized the tech industry. This innovative product catalyzed a flurry of investment in Generative AI.
The innovation journey continued in March 2023 with the introduction of GPT-4\footnote{GPT-4 can accept a prompt of text and images. For clarity, in this paper, we refer to the version of the model with visual capabilities as ``GPT-4V'', following the OpenAI official report~\cite{gpt4v}.}, a large multimodal model, capable of processing both text and images, further captivated the industry by demonstrating the extensive capabilities of multimodal technologies.
By the end of September 2023, GPT-4 with Vision (GPT-4V) was fully integrated into the ChatGPT platform. Following this milestone, comprehensive user study reports~\cite{yang2023dawn,lin2023mm,shi2023exploring,yang2023performance,zhou2023exploring,wen2023road,li2023comprehensive} by computer vision researchers began to emerge, providing evaluations of its visual prowess.
More recently, on the first anniversary of ChatGPT, November 6, 2023, OpenAI hosted its first DevDay, during which the GPT-4V API was released. This release opens new doors for the academic community to conduct extensive evaluations of its performance across a range of visual benchmarks, offering quantitative metrics beyond the limited scope of user studies.

\begin{figure*}[t]
  \centering
   \includegraphics[width=1\linewidth]{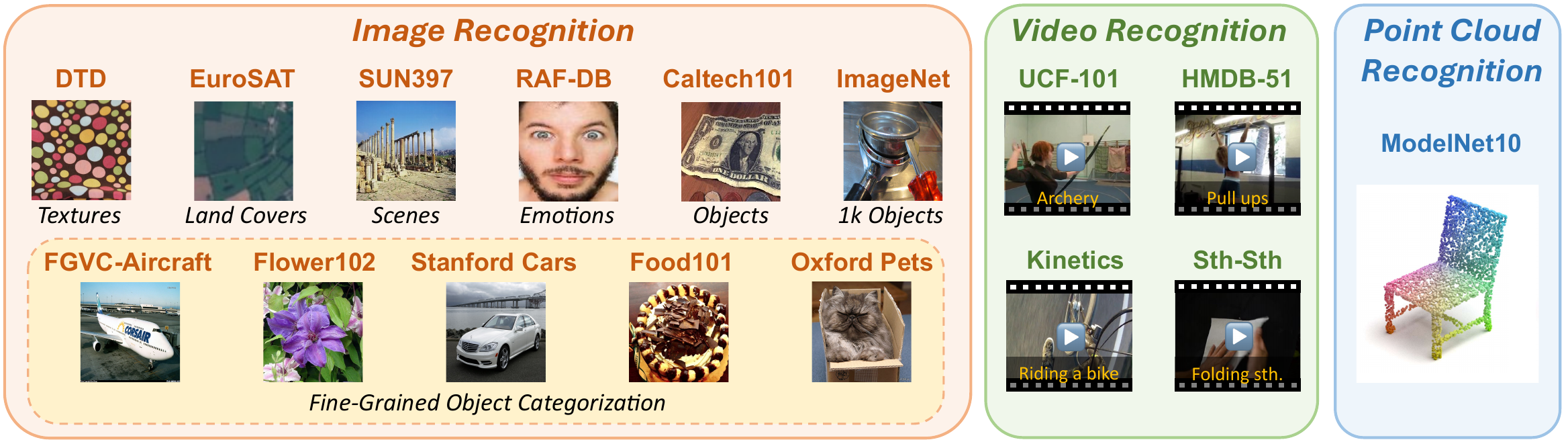}
   \caption{An overview of 16 evaluated popular benchmark datasets, comprising images, videos, and point clouds. The image benchmarks include tasks such as texture recognition, satellite image classification, scene recognition, facial expression recognition, as well as fine-grained object classification. The video datasets encompass diverse human actions captured from various viewpoints and scenes. The point cloud datasets provide valuable information that can be projected onto multi-view depth maps for visual recognition.}
   \label{fig:datasets}
\end{figure*}

In this paper, we evaluate GPT-4's performance in visual recognition tasks—one of the fundamental tasks in the field of computer vision—without any fine-tuning (\ie, in a zero-shot manner). We explore two main facets: \textbf{\whblue{linguistic}} and \textbf{\whred{visual}} capabilities. 
(i) Regarding linguistic capabilities, we investigate how GPT-4's language proficiency can bolster visual recognition. It's widely recognized that the large-scale image-text pre-trained model CLIP~\cite{clip} has built a bridge between vision and text, enabling zero-shot visual recognition by calculating similarity scores between category name embeddings and image embeddings. Building on CLIP's foundation, we aim to utilize GPT-4's extensive linguistic knowledge to craft more nuanced and detailed descriptions of category names, thus enhancing intra-class diversity and inter-class distinguishability, offering a refined alternative to the use of basic category names for zero-shot recognition.
(ii) As for visual capabilities, the evaluation is quite straightforward: we directly feed the image or images (applicable to video and point cloud data) along with candidate categories. By employing prompts, we instruct GPT-4V to organize the candidate categories by relevance to the visual content, thereby obtaining Top-5 prediction.


To conduct a comprehensive evaluation, we included three distinct modalities: images, videos, and point clouds, across 16 well-known and publicly available classification benchmarks~\cite{dtd,helber2019eurosat,xiao2010sun,li2017reliable,caltech101,deng2009imagenet,aircraft,flower,Stanfordcar,bossard2014food,parkhi2012cats,ucf101,hmdb,i3d,sth-sth,modelnet40}, as showcased in Figure~\ref{fig:datasets}. 
For video datasets, we implemented a uniform sampling of frames to create multi-image inputs.
For point cloud data, we process the 3D shape into multiple 2D rendered images.
For each dataset, we offer the zero-shot performance of CLIP, a representative model among web-scale pre-trained vision-language models (VLMs), as a reference. The results includes four backbones: OpenAI CLIP (pre-trained on 400M image-text pairs) with ViT-B/32, ViT-B/16, and ViT-L/14, as well as the more recent and larger EVA CLIP's ViT-E/14, which boasts a staggering 4.4B parameters (14$\times$ that of ViT-L), and has been pre-trained on 2B image-text pairs.
Our research highlights that GPT-4's linguistic capabilities play a crucial role in boosting zero-shot visual recognition, delivering a notable average increase of 7\% in top-1 absolute accuracy across 16 datasets. This leap in performance is driven by GPT-4's rich knowledge, enabling it to craft detailed descriptions for diverse categories. Simultaneously, GPT-4's prowess in visual recognition, evaluated across 16 datasets, is almost on par with EVA-CLIP's ViT-E. This is particularly evident in video datasets such as UCF-101 and HMDB-51, where GPT-4 distinctly surpasses the performance of ViT-E, highlighting its effectiveness in handling contextual visual content reasoning.
For more detailed results, analyses, and experimental details, please refer to the experimental section.

To the best of our knowledge, this study is the first quantitative evaluation of zero-shot visual recognition capabilities using GPT-4V across three modalities—images, videos, and point clouds—over 16 popular visual benchmarks. We believe that the empirical evidence and prompts provided herein are worth knowing. We hope our data points and experience will contribute meaningfully to future research.

\begin{figure*}[t]
  \centering
   \includegraphics[width=0.98\linewidth]{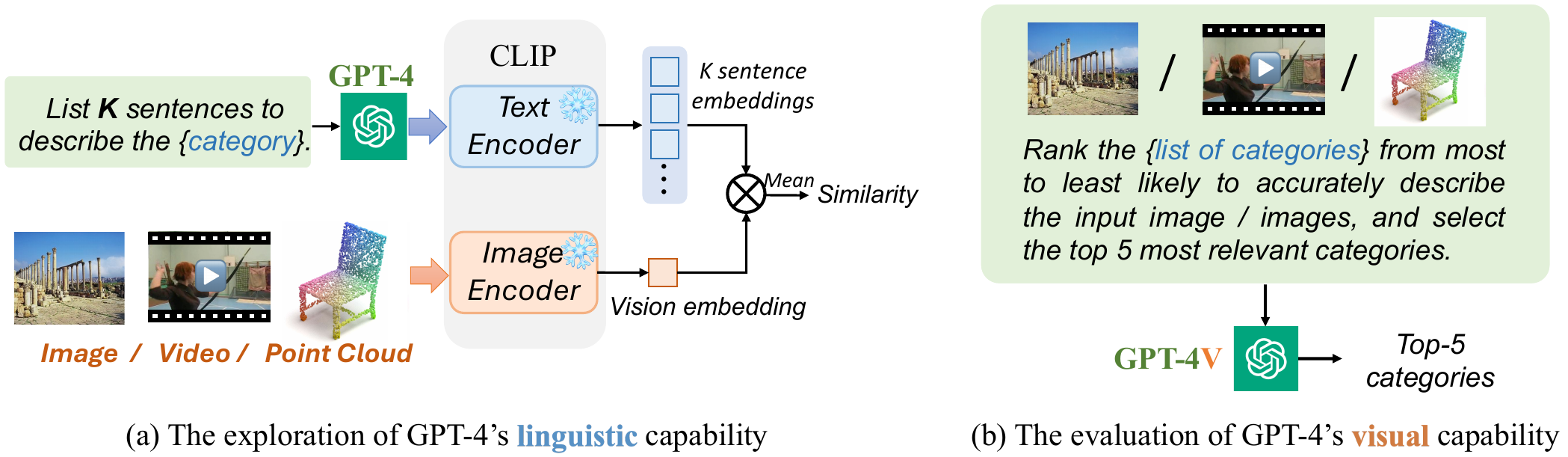}
   \caption{Zero-shot visual recognition leveraging GPT-4's linguistic and visual capabilities. (a) We built upon the visual-language bridge established by CLIP~\cite{clip} and employed the rich linguistic knowledge of GPT-4 to generate additional descriptions for categories, exploring the benefits to visual recognition. (b) We present visual content (\ie, single or multiple images) along with a category list, and prompt GPT-4V to generate the top-5 prediction results.}
   \label{fig:method}
\end{figure*}

%% file: sec/5_related.tex
\section{Related Works}

\noindent\textbf{Preliminary Explorations of GPT-4V.} Recent studies have undertaken detailed case studies on GPT-4V's capabilities across diverse tasks. 
Prior research~\cite{yang2023dawn} delved into the reasoning skills of foundational models within visual domains from a qualitative perspective. 
Subsequently, GPT-4V's performance has been examined in various
visual-language tasks, including but not limited to video understanding~\cite{lin2023mm}, optical character recognition (OCR)~\cite{shi2023exploring}, image context reasoning~\cite{liu2023hallusionbench}, recommender system~\cite{zhou2023exploring}, mathematical logic~\cite{lu2023mathvista}, medical imaging analysis~\cite{yang2023performance,li2023comprehensive}, anomaly detection~\cite{cao2023towards}, social media analysis~\cite{lyu2023gpt} and autonomous driving~\cite{wen2023road}.
However, a gap remains in these studies: most have concentrated on qualitative, initial explorations without extensive quantitative analysis utilizing established visual benchmarks. Such analysis is essential for a comprehensive validation of GPT-4V's visual understanding capabilities. 
The recent availability of its API \footnote{gpt-4-vision-preview: https://platform.openai.com/docs/guides/vision} now enables large-scale quantitative evaluations.

\vspace{1mm}
\noindent\textbf{Enhancing Zero-shot Visual Recognition with LLMs.} The web-scale image-text pre-training model, CLIP~\cite{clip}, has established a pivotal connection between visual and textual domains. Numerous subsequent studies have extended this model to video understanding~\cite{text4vis,wu2023transferring,bike,wu2022cap4video,fang2023uatvr,wang2021actionclip,luo2022clip4clip} or point cloud recognition~\cite{zhang2021pointclip,Zhu2022PointCLIPV2}.
With the ascent of Large Language Models (LLMs), there is increasing interest in harnessing class-specific knowledge from LLMs can improve CLIP’s prediction accuracy.
\cite{menon2022visual} leveraged GPT-3~\cite{gpt3} to create text descriptions for unseen class labels and compare the image embedding with the embeddings of the descriptions. \cite{ren2023chatgpt} further developed this concept by employing ChatGPT to structure the classes hierarchically to boost zero-shot image recognition.
In our study, rather than constructing a hierarchical structure, we prompt GPT-4 to produce multi-sentence descriptions for categories, examining the effectiveness of this straightforward approach in image, video, and point cloud recognition.

%% file: sec/2_apporach.tex
\section{Methodology}
As demonstrated in Figure~\ref{fig:method}, we evaluate GPT-4's linguistic and visual capabilities in zero-shot visual recognition. This section will introduce the specific details.

\subsection{Data Processing}

\begin{figure}[h]
  \centering
   \includegraphics[width=0.98\linewidth]{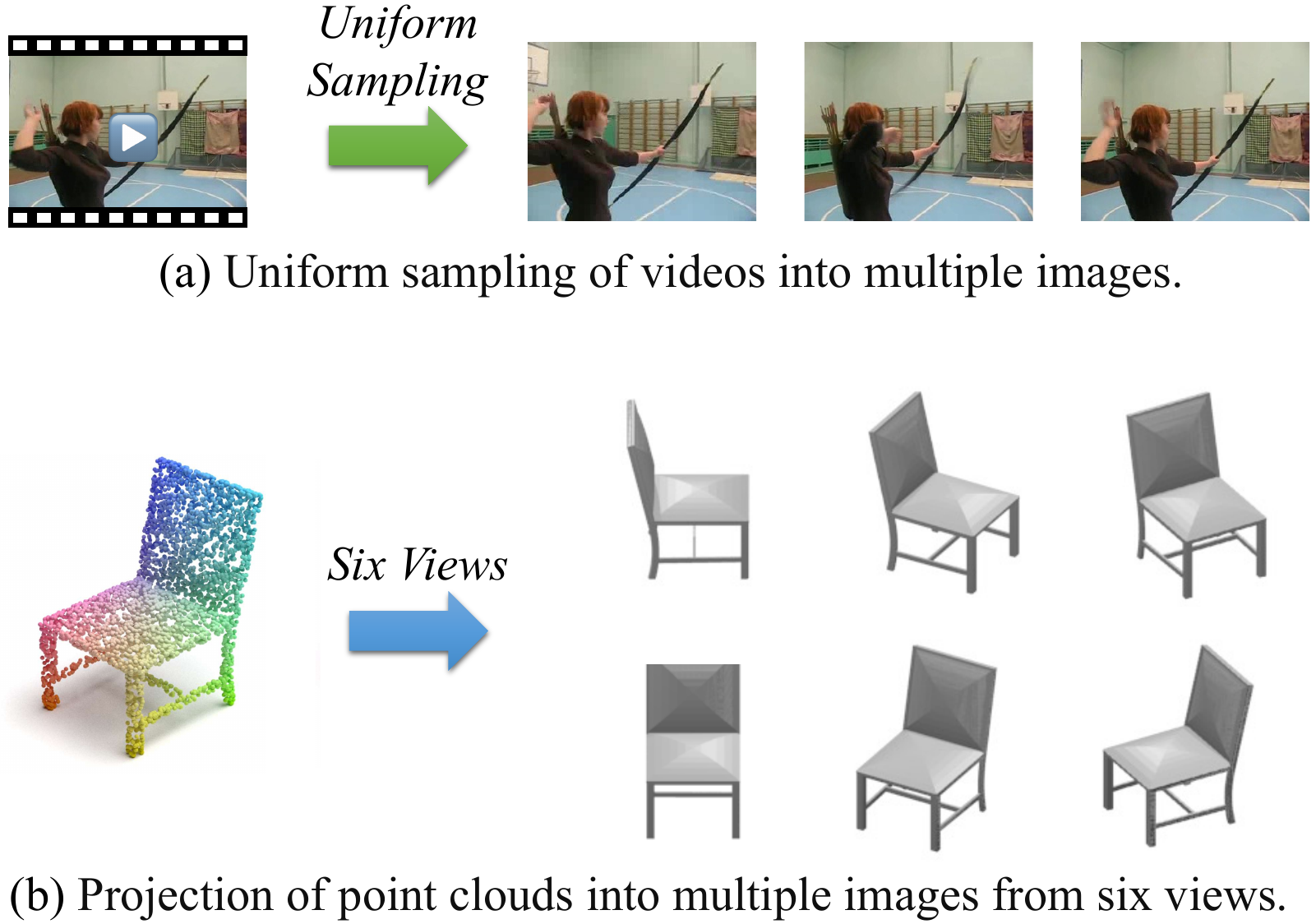}
   \caption{Processing video and point cloud data into images.}
   \label{fig:data_process}
\end{figure}

To align with the input requirements of CLIP~\cite{clip} and GPT-4V~\cite{gpt4v}, apart from image classification tasks, the inputs for video and point cloud classification must be transformed into images. 
As illustrated in Figure~\ref{fig:data_process}, this process involves the transformation of video and point cloud data into image sets. Specifically, for input videos, we uniformly sample eight frames to serve as multi-image inputs. Regarding point clouds, we follow MVCNN~\cite{su2015multi} to render multiple views around the object in a uni-directional manner at an angle of 30 degrees. To reduce testing costs, we use six frontally rendered images in our process.
This prepares us to carry out subsequent evaluations.

\subsection{Exploration of Linguistic Capabilities}
Our objective is to explore how the extensive linguistic knowledge of GPT-4 can be leveraged to enhance visual recognition performance. Building on the cross-modal bridge established by CLIP through large-scale image-text pre-training, we aim to enrich textual descriptions beyond using simple category names to better align with visual content. As shown in Figure~\ref{fig:method}(a), we begin by guiding GPT-4 to generate $K$ sentences describing each category in the dataset using appropriate prompts. These $K$ sentences are then converted into $K$ text embeddings via CLIP's frozen text encoder, while the visual signal is encoded into a vision embedding by CLIP's frozen image encoder (\eg, for video and point cloud data, the vision embedding is obtained by global averaging pooling over multiple frame embeddings or viewpoint embeddings). Subsequently, these text embeddings are compared with the vision embedding to calculate $K$ similarity scores. After normalization with a \emph{Softmax} function and averaging, we obtain a consolidated similarity score for each category in relation to the visual input.  Given a dataset with $C$ categories, each visual input yields $C$ similarity scores, which are then ranked from highest to lowest to determine the final prediction.

\subsection{Evaluation of Visual Capabilities}
The recent release of the GPT-4V API allows for a comprehensive evaluation of visual benchmarks, moving beyond limited case studies within the ChatGPT web interface. The evaluation process, outlined in Figure~\ref{fig:method}(b), is simple and straightforward. Visual samples, whether a single image or a collection, are inputted along with an appropriate text prompt. This prompt guides GPT-4V to assess the dataset's categories, sorting them based on their relevance to the provided visual content, and produces the top-5 prediction results. These predictions are subsequently compared against the ground truth of the dataset to derive both top-1 and top-5 accuracy metrics, providing a quantitative assessment of GPT-4V's visual understanding capabilities. 
Further details about our prompts is available in \href{https://github.com/whwu95/GPT4Vis}{Code Repo}.

%% file: sec/3_experiment.tex
\section{Experiments}

\subsection{Datasets}

\begin{table}[t]
    \centering
    \scalebox{0.83}{
    \setlength{\tabcolsep}{2.0pt}
    \begin{tabular}{llccc} 
    \toprule
     Modality &  Dataset & \#Classes & Res. (Ave.) & \#Samples \\ \midrule
     \multirow{11}{*}{Image} &  DTD~\cite{dtd} & 47 & 300\x300 &  1,692  \\
       &  EuroSAT~\cite{helber2019eurosat} & 10 & 64\x64 & 8,100 \\
       & SUN397~\cite{xiao2010sun} &  397 & 969\x776   & 19,850 \\
       & RAF-DB~\cite{li2017reliable} & 7 & 100\x100 &  3,068 \\
       & Caltech101~\cite{caltech101} & 101 & 300\x200  & 2,465 \\
       & ImageNet~\cite{deng2009imagenet} &  1000 & 469\x387  & 50,000 \\ 
       & FGVC-Aircraft~\cite{aircraft} & 100 & 1098\x747 & 3,333 \\
       & Flower102~\cite{flower} & 102 & 667\x500 & 2,463 \\
       & Stanford Cars~\cite{Stanfordcar} & 196 & 360\x240 & 8,041 \\
       & Food101~\cite{bossard2014food} &  101 & 500\x350 & 30,300 \\
       & Oxford Pets~\cite{parkhi2012cats} & 37 & 500\x350 & 3,669 \\ \midrule
     \multirow{4}{*}{Video}  & UCF-101~\cite{ucf101} (Split 1) & 101 & 320\x240 & 3,783 \\
       & HMDB-51~\cite{hmdb} (Split 1) & 51 & 340\x256 & 1,530 \\
       & Kinetics-400~\cite{i3d} & 400 & 420\x320 & 19,796 \\
       & Sth-Sth V1~\cite{sth-sth} & 174 & 176\x100 & 11,522 \\ \midrule
    Point Cloud   & ModelNet10~\cite{modelnet40} & 10 & 224\x224 & 908 \\
    \bottomrule
    \end{tabular}
    }
    \caption{The statistics of these evaluated datasets.}
    \label{tab:datasets}
\end{table}

This study evaluates 16 visual datasets across images, videos, and point clouds. The evaluation employs the widely recognized validation sets for these benchmarks, with Table~\ref{tab:datasets} providing detailed statistics for each dataset.

\subsection{Implementation Details}

\begin{figure}[!t]
  \centering
   \includegraphics[width=0.985\linewidth]{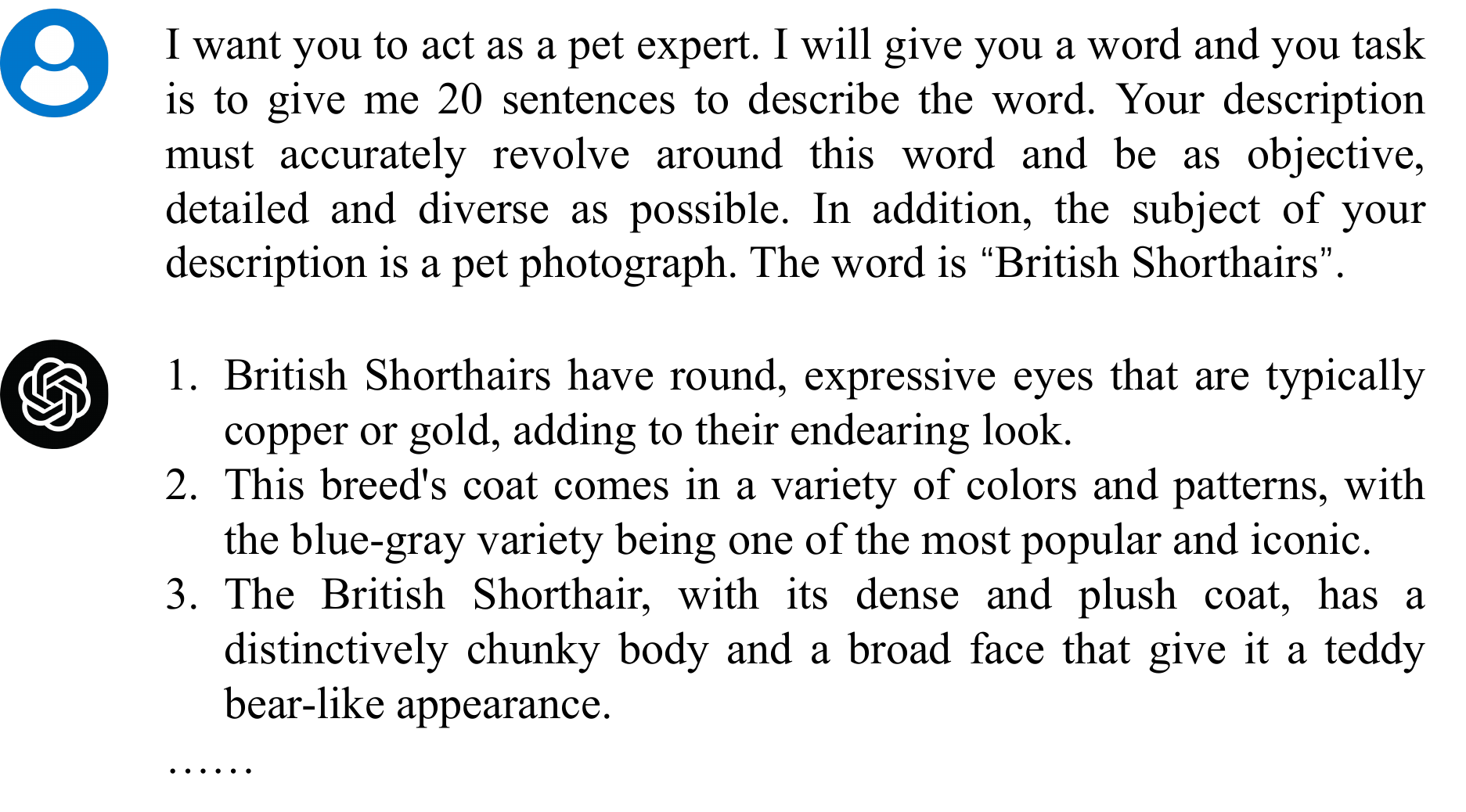}
   \caption{Sentences generated by GPT-4 for ``British Shorthairs''.}
   \label{fig:gpt4_sentences}
\end{figure}

\begin{figure}[!t]
  \centering
   \includegraphics[width=0.985\linewidth]{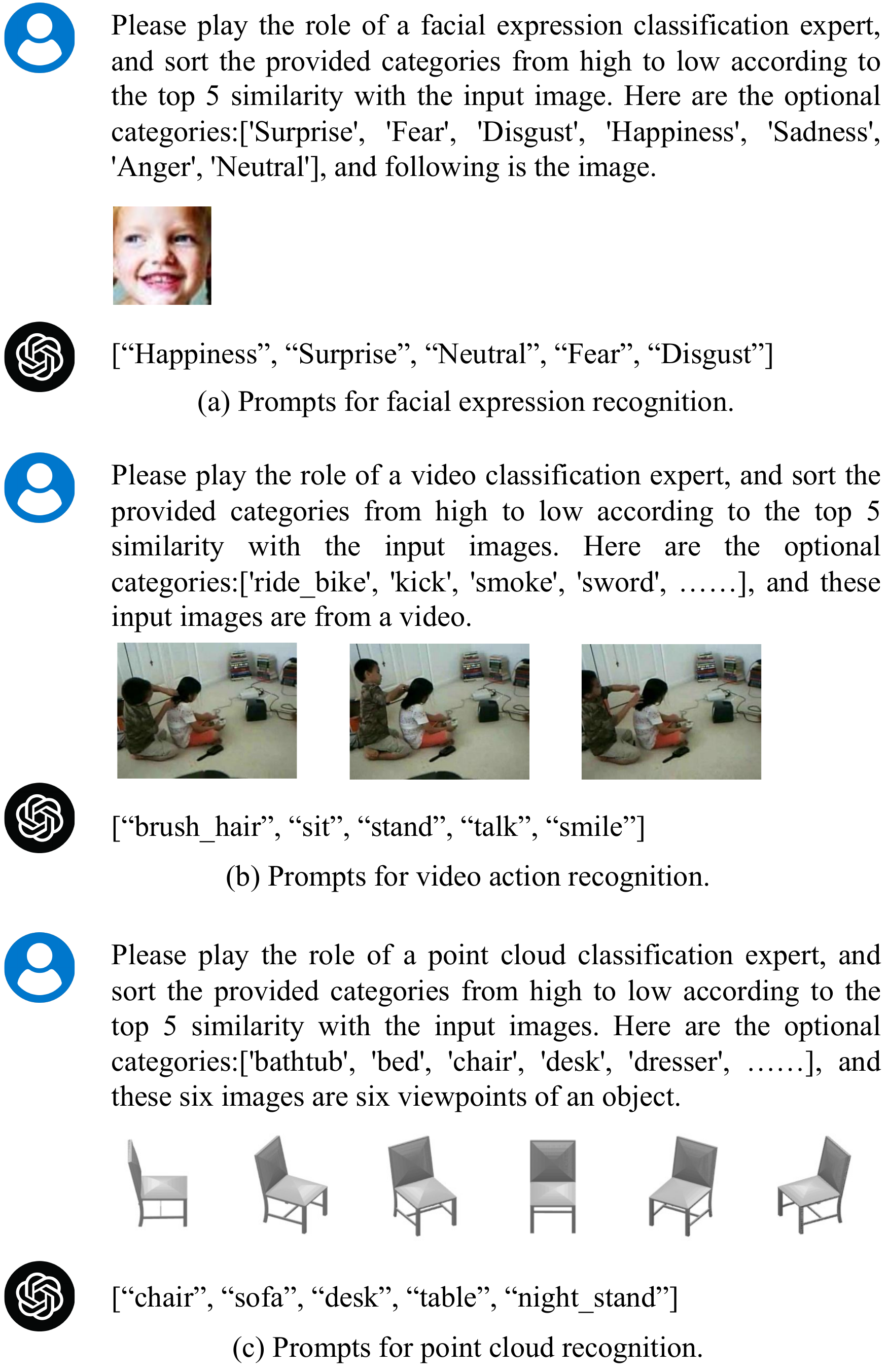}
   \caption{Prompts for image, video, and point cloud datasets: (a) An example from RAF-DB~\cite{li2017reliable} illustrates 7-class facial expression recognition. (b) A video example from HMDB-51~\cite{hmdb} demonstrates 51-class action recognition, where ellipses indicate category names omitted due to space constraints. (c) An example from ModelNet10~\cite{modelnet40} for point cloud classification across 10 categories, where ellipses again indicate the truncation of category names owing to space constraints. Please zoom in for best view.} 
   \label{fig:gpt4v}
\end{figure}

\noindent\textbf{GPT-4 Generated Descriptions.}
Using the GPT-4 API (version gpt-4-1106-preview), we generate $K$ descriptive sentences for each category, with $K$ defaulting to be 20. As an example, for the ``British Shorthairs'' category from the Oxford Pets dataset~\cite{parkhi2012cats}, we present our prompts alongside GPT-4's responses in Figure~\ref{fig:gpt4_sentences}.

\vspace{1mm}
\noindent\textbf{Utilizing GPT-4 with Vision.}
In our study, we employ the GPT-4V API (specifically, gpt-4-vision-preview) to evaluate 16 different benchmarks. For videos, we select 8 frames via uniform sampling for API processing, and for point clouds, we provide images from six perspectives. Figure~\ref{fig:gpt4v} showcases the interaction with GPT-4V, highlighting both the prompts used and the subsequent responses for evaluations across images, videos, and point clouds.

\begin{table*}[t]
    \centering
    \resizebox{\textwidth}{!}{
    \begin{NiceTabular}{l*{9}c}
        \toprule
        Dataset & \Block{1-3}{Describable Textures (DTD)} & & & \Block{1-3}{EuroSAT} & & & \Block{1-3}{SUN397} & & \\
        Backbone (\#Param) & Baseline & \textbf{\whblue{GPT Prompts}} & Top-1 \(\Delta\) & Baseline & \textbf{\whblue{GPT Prompts}} & Top-1 \(\Delta\) & Baseline & \textbf{\whblue{GPT Prompts}} & Top-1 \(\Delta\) \\ \cmidrule(lr){1-1} \cmidrule(lr){2-4} \cmidrule(lr){5-7} \cmidrule(lr){8-10}
        CLIP ViT-B/32 (88M) & 42.1 / 69.3 & \gpt{46.9 / 78.0} & \textcolor{teal}{+4.8} & 40.2 / 87.4 & \gpt{49.4 / 83.1} & \textcolor{teal}{+9.2} & 59.2 / 88.1 & \gpt{63.3 / 91.0} & \textcolor{teal}{+4.1} \\ 
        CLIP ViT-B/16 (86M) & 46.0 / 72.6 & \gpt{48.5 / 78.4} & \textcolor{teal}{+2.5} & 45.8 / 81.0 & \gpt{48.8 / 83.7} & \textcolor{teal}{+3.0} & 60.9 / 89.1 & \gpt{65.5 / 91.9} & \textcolor{teal}{+4.6} \\
        CLIP ViT-L/14 (304M) & 51.8 / 77.4 & \gpt{54.8 / 81.6} & \textcolor{teal}{+3.0} & 44.0 / 95.2 & \gpt{54.1 / 95.0} & \textcolor{teal}{+10.1} & 65.2 / 91.4 & \gpt{70.3 / 94.2} & \textcolor{teal}{+5.1} \\
        EVA ViT-E/14 (4.4B) & 61.5 / 86.5 & \gpt{66.5 / 94.7} & \textcolor{teal}{+5.0} & 55.1 / 96.9 & \gpt{70.3 / 93.7} & \textcolor{teal}{+15.2} & 71.6 / 92.3 & \gpt{75.6 / 95.8} & \textcolor{teal}{+4.0} \\ \midrule
        \rpink 
        \textbf{\whred{GPT-4V}} & \Block{1-3}{ 57.7 / 83.3} & & & \Block{1-3}{46.8 / 86.2}  & &  & \Block{1-3}{59.2 / 78.1} & & \\
        \bottomrule
        \toprule
        Dataset & \Block{1-3}{Real-world Affective Faces (RAF-DB)} & & & \Block{1-3}{Caltech101} & & & \Block{1-3}{ImageNet-1K} & & \\
        Backbone (\#Param)  & Baseline & \textbf{\whblue{GPT Prompts}} & Top-1 \(\Delta\) & Baseline & \textbf{\whblue{GPT Prompts}} & Top-1 \(\Delta\)  & Baseline & \textbf{\whblue{GPT Prompts}} & Top-1 \(\Delta\)  \\ \cmidrule(lr){1-1} \cmidrule(lr){2-4} \cmidrule(lr){5-7} \cmidrule(lr){8-10}
        CLIP ViT-B/32 (88M) & 22.4 / 76.6 & \gpt{45.8 / 90.6} & \textcolor{teal}{+23.4} & 86.8 / 99.1 & \gpt{92.8 / 99.6} & \textcolor{teal}{+6.0} & 59.0 / 85.6 & \gpt{63.7 / 88.7} & \textcolor{teal}{+4.7} \\ 
        CLIP ViT-B/16 (86M) & 27.5 / 69.1 & \gpt{54.4 / 94.4} & \textcolor{teal}{+26.9} & 87.9 / 98.7 & \gpt{94.6 / 99.6} & \textcolor{teal}{+6.7} & 64.1 / 89.3 & \gpt{68.7 / 91.6} & \textcolor{teal}{+4.6} \\
        CLIP ViT-L/14 (304M) & 26.1 / 72.1 & \gpt{47.2 / 92.0} & \textcolor{teal}{+21.1} & 86.7 / 99.3 & \gpt{96.2 / 100.0} & \textcolor{teal}{+9.5} & 71.6 / 92.2 & \gpt{75.5 / 94.5} & \textcolor{teal}{+4.9} \\
        EVA ViT-E/14 (4.4B) & 31.0 / 90.9 & \gpt{54.9 / 93.7} & \textcolor{teal}{+23.9} & 94.0 / 99.7 & \gpt{97.9 / 100.0} & \textcolor{teal}{+3.9} & 78.4 / 93.5 & \gpt{81.6 / 96.1} & \textcolor{teal}{+3.2} \\ \midrule
        \rpink
        \textbf{\whred{GPT-4V}} & \Block{1-3}{68.7 / 93.8} & & & \Block{1-3}{93.7 / 98.2}  & & & \Block{1-3}{63.1 / 78.2}  & &  \\
        \bottomrule
        \toprule
        Dataset & \Block{1-3}{FGVC-Aircraft} & & & \Block{1-3}{Flower102} & & & \Block{1-3}{Stanford Cars} & & \\
        Backbone (\#Param)  & Baseline & \textbf{\whblue{GPT Prompts}} & Top-1 \(\Delta\) & Baseline & \textbf{\whblue{GPT Prompts}} & Top-1 \(\Delta\)  & Baseline & \textbf{\whblue{GPT Prompts}} & Top-1 \(\Delta\)  \\ \cmidrule(lr){1-1} \cmidrule(lr){2-4} \cmidrule(lr){5-7} \cmidrule(lr){8-10}
        CLIP ViT-B/32 (88M) & 16.6 / 44.3 & \gpt{21.9 / 54.3} & \textcolor{teal}{+5.3} & 61.6 / 77.6 & \gpt{71.8 / 89.6} & \textcolor{teal}{+10.2} & 58.9 / 90.8 & \gpt{61.2 / 92.5} & \textcolor{teal}{+2.3} \\ 
        CLIP ViT-B/16 (86M) & 21.1 / 55.0 & \gpt{28.0 / 65.4} & \textcolor{teal}{+6.9} & 64.8 / 80.2 & \gpt{74.5 / 91.4} & \textcolor{teal}{+9.7} & 63.6 / 93.7 & \gpt{66.8 / 95.6} & \textcolor{teal}{+3.2} \\
        CLIP ViT-L/14 (304M) & 27.3 / 69.5 & \gpt{36.3 / 82.2} & \textcolor{teal}{+9.0} & 73.1 / 86.3 & \gpt{81.5 / 94.0} & \textcolor{teal}{+8.4} & 76.2 / 97.9 & \gpt{77.9 / 98.6} & \textcolor{teal}{+1.7} \\
        EVA ViT-E/14 (4.4B) & 50.6 / 86.9 & \gpt{58.4 / 96.6} & \textcolor{teal}{+7.8} & 82.1 / 90.8 & \gpt{87.0 / 95.0} & \textcolor{teal}{+4.9} & 94.2 / 99.8 & \gpt{94.5 / 99.9} & \textcolor{teal}{+0.3} \\ \midrule
        \rpink 
        \textbf{\whred{GPT-4V}} & \Block{1-3}{56.6 / 80.8} & & & \Block{1-3}{69.1 / 77.3}  & & & \Block{1-3}{62.7 / 81.8}  & &  \\
        \bottomrule
        \toprule
       Dataset & \Block{1-3}{Food101} & & & \Block{1-3}{Oxford Pets} & & & \Block{1-3}{UCF-101} & & \\
        Backbone (\#Param)  & Baseline & \textbf{\whblue{GPT Prompts}} & Top-1 \(\Delta\) & Baseline & \textbf{\whblue{GPT Prompts}} & Top-1 \(\Delta\)  & Baseline & \textbf{\whblue{GPT Prompts}} & Top-1 \(\Delta\)  \\ \cmidrule(lr){1-1} \cmidrule(lr){2-4} \cmidrule(lr){5-7} \cmidrule(lr){8-10}
        CLIP ViT-B/32 (88M) & 78.0 / 95.1 & \gpt{80.8 / 96.1} & \textcolor{teal}{+2.8} & 79.9 / 96.3 & \gpt{89.3 / 99.7} & \textcolor{teal}{+9.4} & 59.9 / 83.1 & \gpt{69.9 / 93.1} & \textcolor{teal}{+10.0} \\ 
        CLIP ViT-B/16 (86M) & 84.0 / 97.1 & \gpt{86.3 / 97.8} & \textcolor{teal}{+2.3} & 81.8 / 96.4 & \gpt{91.0 / 99.8} & \textcolor{teal}{+9.2} & 64.4 / 84.8 & \gpt{72.0 / 93.4} & \textcolor{teal}{+7.6} \\
        CLIP ViT-L/14 (304M) & 89.9 / 98.5 & \gpt{91.4 / 98.7} & \textcolor{teal}{+1.5} & 88.2 / 97.7 & \gpt{94.1 / 99.9} & \textcolor{teal}{+5.9} & 72.3 / 92.4 & \gpt{80.6 / 97.0} & \textcolor{teal}{+8.3} \\
        EVA ViT-E/14 (4.4B) & 93.4 / 99.0 & \gpt{93.2 / 99.1} & \textcolor{blue}{-0.2} & 93.0 / 98.7 & \gpt{95.8 / 99.9} & \textcolor{teal}{+2.8} & 74.8 / 93.3 & \gpt{86.5 / 99.0} & \textcolor{teal}{+11.7} \\ \midrule
        \rpink
        \textbf{\whred{GPT-4V}} & \Block{1-3}{86.2 / 93.8} & & & \Block{1-3}{90.8 / 98.6}  & & & \Block{1-3}{83.7 / 94.9} & &  \\
        \bottomrule
        \toprule
       Dataset & \Block{1-3}{HMDB-51} & & & \Block{1-3}{Kinetics-400} & & & \Block{1-3}{Something-Something V1} & & \\
        Backbone (\#Param)  & Baseline & \textbf{\whblue{GPT Prompts}} & Top-1 \(\Delta\) & Baseline & \textbf{\whblue{GPT Prompts}} & Top-1 \(\Delta\)  & Baseline & \textbf{\whblue{GPT Prompts}} & Top-1 \(\Delta\)  \\ \cmidrule(lr){1-1} \cmidrule(lr){2-4} \cmidrule(lr){5-7} \cmidrule(lr){8-10}
        CLIP ViT-B/32 (88M) & 38.4 / 64.9 & \gpt{47.2 / 78.4} & \textcolor{teal}{+8.8} & 47.9 / 75.2 & \gpt{52.7 / 79.5} & \textcolor{teal}{+4.8} & 2.2 / 7.9 & \gpt{3.0 / 10.6} & \textcolor{teal}{+0.8} \\ 
        CLIP ViT-B/16 (86M) & 41.9 / 69.9 & \gpt{51.1 / 80.5} & \textcolor{teal}{+9.2} & 52.8 / 78.4 & \gpt{55.2 / 82.1} & \textcolor{teal}{+2.4} & 2.7 / 9.2 & \gpt{3.4 / 11.2} & \textcolor{teal}{+0.7} \\
        CLIP ViT-L/14 (304M) & 46.8 / 75.1 & \gpt{55.6 / 86.0} & \textcolor{teal}{+8.8} & 60.6 / 83.9 & \gpt{63.3 / 87.2} & \textcolor{teal}{+2.7} & 3.7 / 11.9 & \gpt{3.9 / 13.8} & \textcolor{teal}{+0.2} \\
        EVA ViT-E/14 (4.4B) & 41.5 / 70.3 & \gpt{56.4 / 86.7} & \textcolor{teal}{+14.9} & 61.5 / 84.9 & \gpt{67.6 / 87.7} & \textcolor{teal}{+6.1} & 3.8 / 12.2 & \gpt{5.3 / 16.2} & \textcolor{teal}{+1.5} \\ \midrule
        \rpink
        \textbf{\whred{GPT-4V}} & \Block{1-3}{63.2 / 86.9} & & & \Block{1-3}{58.8 / 75.7}  & & & \Block{1-3}{4.6 / 11.3}  & &  \\
        \bottomrule
        \ctoprule{1-7}
       Dataset & \Block{1-3}{ModelNet10} & & & \Block{1-3}{\textbf{Average over 16 datasets}} \\
        Backbone (\#Param)  & Baseline & \textbf{\whblue{GPT Prompts}} & Top-1 \(\Delta\)  & Baseline & \textbf{\whblue{GPT Prompts}} & Top-1 \(\Delta\)  \\ \cmidrule(lr){1-1} \cmidrule(lr){2-4} \cmidrule(lr){5-7}
        CLIP ViT-B/32 (88M) & 42.9 / 84.8 & \gpt{47.8 / 86.1} & \textcolor{teal}{+4.9} & 49.7 / 76.6 & \gpt{56.7 / 81.9} & \textcolor{teal}{+7.0} \\ 
        CLIP ViT-B/16 (86M) & 39.1 / 82.2 & \gpt{48.7 / 93.3} & \textcolor{teal}{+9.6}  & 53.0 / 77.9 & \gpt{59.8 / 84.4} & \textcolor{teal}{+6.8} \\
        CLIP ViT-L/14 (304M) & 59.5 / 88.1 & \gpt{60.2 / 92.2} & \textcolor{teal}{+0.7} & 58.9 / 83.1 & \gpt{65.2 / 87.9} & \textcolor{teal}{+6.3} \\
        EVA ViT-E/14 (4.4B) & 70.0 / 94.2 & \gpt{80.3 / 99.7} & \textcolor{teal}{+10.3} & 66.0 / 86.9 & \gpt{73.2 / 90.9} & \textcolor{teal}{+7.2} \\ 
        \cmidrule{1-7}
        \rpink
        \textbf{\whred{GPT-4V}} & \Block{1-3}{66.8 / 90.9} & & & \Block{1-3}{64.5 / 81.9}  & & \\
        \cbottomrule{1-7}
    \end{NiceTabular}
    }
    \caption{Main results in zero-shot visual recognition across the 16 datasets, reporting Top-1 and Top-5 accuracy (\%). We also report the parameter count of CLIP's image backbone for reference. ``Baseline'' denotes the direct use of category names. ``\textbf{\whblue{GPT Prompts}}'' refers to the utilization of multi-sentence descriptions generated by GPT-4 Turbo API for category names. ``\textbf{\whred{GPT-4V}}'' indicates the use of the GPT-4 Turbo with vision API for visual content recognition.}
    \label{tab: gain} 
\end{table*}



\subsection{Gains from \whblue{GPT Prompts}}
Table~\ref{tab: gain} showcases our evaluation results on 16 datasets and their average performance. 
For each dataset, we've detailed results using four different CLIP backbones, including OpenAI CLIP~\cite{clip}'s configurations of ViT-B/32, ViT-B/16, and ViT-L/14, each pre-trained with 400 million image-text pairs, and the EVA CLIP~\cite{sun2023eva}'s ViT-E/14, which is notable for its 4.4 billion parameters (14$\times$ that of ViT-L/14) and training on 2 billion image-text pairs. We will delve into an analysis of these results next.

Descriptions generated by GPT-4 distinctly surpass the CLIP baseline in a majority of datasets, boasting an average top-1 accuracy improvement of 7\% across 16 datasets. This consistent enhancement across all three modalities—images, videos, and point clouds—highlights the method's potent generalizability. More specifically:  

1) For image datasets, with RAF-DB~\cite{li2017reliable} as a focal point, GPT Prompts enable an over 20\% increment in accuracy across various backbones. For other datasets like EuroSAT~\cite{helber2019eurosat} satellite image classification, Flower~\cite{flower} fine-grained recognition, Pets~\cite{parkhi2012cats} fine-grained recognition, Aircraft~\cite{aircraft} fine-grained classification, and Caltech101~\cite{caltech101} object classification, we observe boosts of approximately 9-15\%. Smaller gains in Stanford Cars~\cite{Stanfordcar} and Food101~\cite{bossard2014food} suggest that a high density of similar categories may lead to ambiguous descriptions, confusing the CLIP model. In general, larger CLIP models achieve better zero-shot recognition performance on image tasks, and GPT-generated prompts reliably offer additional enhancements.

\begin{figure}[t]
  \centering
  \begin{subfigure}{1\linewidth}
    \includegraphics[width=\textwidth]{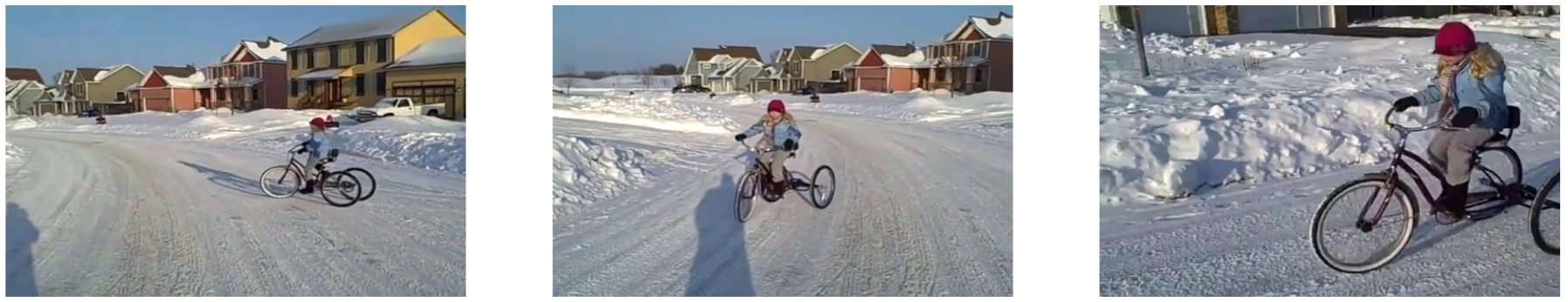}
    \caption{Ground Truth: Biking through snow. }
    \label{fig:k400_case1}
  \end{subfigure}
  \begin{subfigure}{1\linewidth}
    \includegraphics[width=\textwidth]{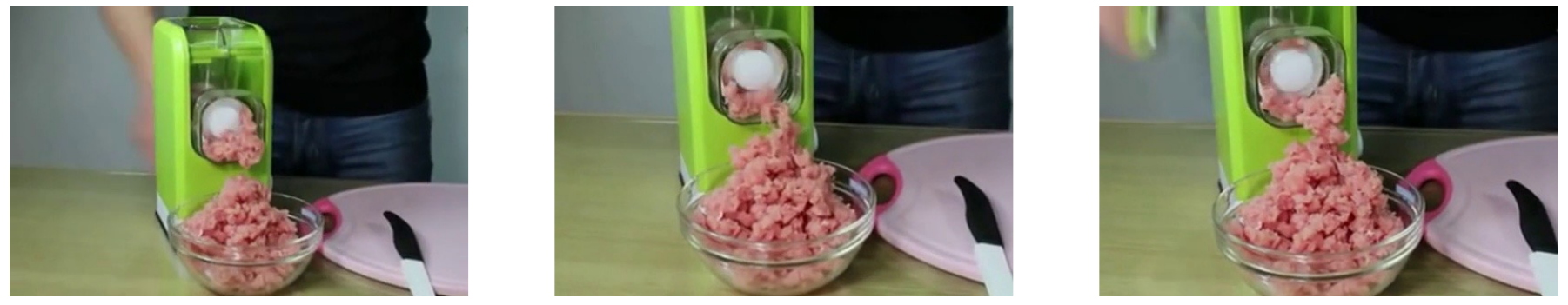}
    \caption{Ground Truth: Grinding meat.}
    \label{fig:k400_case2}
  \end{subfigure}
   \caption{Two video examples from the Kinetics dataset~\cite{i3d} accurately predicted by GPT-4V.}
  \label{fig:k400}
\end{figure}

\begin{figure}[t]
  \centering
  \begin{subfigure}{1\linewidth}
    \includegraphics[width=\textwidth]{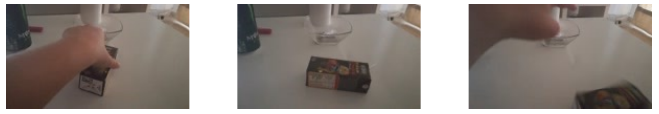}
    \caption{Ground Truth: Pushing something so it spins (\faCheck). GPT-4V Prediction: Pretending to pick something up (\faTimes). }
    \label{fig:case1}
  \end{subfigure}
  \begin{subfigure}{1\linewidth}
    \includegraphics[width=\textwidth]{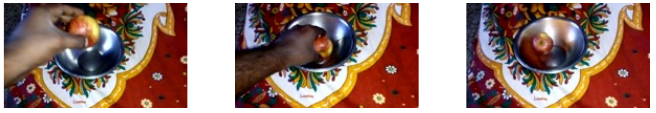}
    \caption{Ground Truth: Putting something into something (\faCheck). GPT-4V Prediction: Opening something (\faTimes).}
    \label{fig:case2}
  \end{subfigure}
   \caption{Two video examples from the Something-Something dataset~\cite{sth-sth} incorrectly predicted by GPT-4V.}
  \label{fig:SSV1}
\end{figure}

2) On video datasets, especially HMDB-51~\cite{hmdb} and UCF101~\cite{ucf101}, we observe astonishing gains of up to 11-15\%, indicating that rich descriptions of human actions align better with video content than simpler phrases. The Something-Something V1 (SSV1)~\cite{sth-sth} dataset, however, exhibits poor performance with the CLIP baseline (\textless~4\% Top-1) due to the lack of temporal modeling. Unlike Kinetics, UCF, and HMDB datasets, which can be recognized through scenes and object appearances as shown in Figure~\ref{fig:k400}, SSV1 demands the understanding of complex object-object and human-object interactions, requiring robust temporal and motion modeling for correct recognition. Hence, activities cannot be inferred merely from individual frames (\eg, Pushing something so it spins), as demonstrated in Figure~\ref{fig:SSV1}. 
In essence, with scene-based video recognition datasets, the larger the CLIP model, the greater the zero-shot performance, a trend consistent with image tasks where GPT Prompts lead to additional gains. Yet, in datasets where temporal modeling is crucial, CLIP's simple frame averaging strategy falls short, and GPT prompts cannot compensate for this deficiency.

3) For point cloud datasets, employing multiple rendered viewpoints for zero-shot recognition with CLIP achieves noteworthy accuracy, mirroring the positive effects seen with image and scene-based video datasets. The integration of GPT Prompts further amplifies these positive results.

\begin{table*}[t]
  \centering
  \scalebox{0.82}{
  \setlength{\tabcolsep}{2.5pt}
    \begin{tabular}{lccccccccccc|ccc|c}
    \toprule
    Prompt for CLIP ViT-B/32 & DTD & SAT & SUN & RAF & Caltech & ImageNet & Aircraft & Flower & Cars & Food & Pets & K400 & UCF & HMDB & MNet10\\
    \midrule
    Baseline: Category name                      & 42.1 & 40.2 & 59.2 & 22.4 & 86.8 & 59.0 & 16.6 & 61.6 & 58.9 & 78.0 & 79.9 & 47.9 & 59.9 & 38.4 & {42.9} \\
    Hand-crafted Prompt                & 43.7 & 45.3 & {62.0} & 24.2 & {91.1} & {62.0} & 19.5 & 67.0 & {60.4} & {80.5} & 87.4 & {49.8} & 61.5 & 43.0 & 40.1 \\
    \whblue{\textbf{GPT Prompts}}                & {44.6} & \gpt{\textbf{49.4}} & 57.7 & \gpt{\textbf{45.8}} & 90.8 & 59.6 & {21.5} & \gpt{\textbf{71.8}} & 53.0 &  80.0 & {87.4} & 47.9 & {69.4} & {44.5} & \gpt{\textbf{47.8}} \\
    Hand-crafted Prompt + \whblue{\textbf{GPT Prompts}} & \gpt{\textbf{46.9}} & {48.0} & \gpt{\textbf{63.3}} & {34.5} & 
    \gpt{\textbf{92.8}} & \gpt{\textbf{63.7}} & \gpt{\textbf{21.9}} & {70.1} & \gpt{\textbf{61.2}} & \gpt{\textbf{80.8}} & \gpt{\textbf{89.3}} & \gpt{\textbf{52.7}} & \gpt{\textbf{69.9}} & \gpt{\textbf{47.2}} & 42.0 \\
    \bottomrule
    \end{tabular}
    }
  \caption{Evaluating the impact of different prompts on CLIP-based zero-shot visual recognition in image, video, and point cloud datasets. ``Hand-crafted Prompt'' denotes a fixed template, such as ``A photo of a \{category name\}." for image datasets, ``A video of a person \{category name\}." for video datasets, and ``A point cloud depth map of a \{category name\}." for point cloud datasets. ``\whblue{\textbf{GPT Prompts}}'' refers to descriptive sentences generated by GPT-4. ``Hand-crafted Prompt + \whblue{\textbf{GPT Prompts}}'' refers to a concatenation of a template with each descriptive sentence generated by GPT-4, such as ``A photo of a \{Category\}. \{GPT-generated sentence\}". 
}
  \label{tab:clip_prompt}
\end{table*}

\subsection{Zero-shot Visual Performance of \whred{GPT-4V}}
To evaluate the visual capabilities of GPT-4V, as shown in Table~\ref{tab: gain}, we conduct quantitative evaluation across 16 datasets. Utilizing straightforward prompts (depicted in Figure~\ref{fig:gpt4v}), we obtain predictions from GPT-4V. 
Analyzing the average results from these 16 datasets, GPT-4V's top-1 accuracy approaches that of EVA's ViT-E. Specifically:

1) On image datasets, GPT-4V significantly outstrips the largest CLIP model EVA ViT-E on the RAF-DB dataset~\cite{li2017reliable} (68.7\% \vs 31.0\%), demonstrating a strong capability in facial expression recognition. Additionally, it outperforms EVA ViT-E in the fine-grained task of aircraft recognition~\cite{aircraft} (56.6\% vs. 50.6\%) and achieves comparable results in Caltech101 object recognition~\cite{caltech101}.
GPT-4V's ability to classify textures~\cite{dtd}, satellite images~\cite{helber2019eurosat}, and recognize pets~\cite{parkhi2012cats} situates it between the performance levels of CLIP ViT-L and EVA ViT-E. However, it slightly lags behind ViT-L in the more specialized areas of flower~\cite{flower} and food~\cite{bossard2014food} recognition. In the broad-spectrum challenge of ImageNet~\cite{deng2009imagenet} 1k-class recognition and car type identification~\cite{Stanfordcar}, GPT-4V's accuracy falls between that of ViT-B/32 and ViT-B/16.
In scenarios such as scene recognition~\cite{xiao2010sun}, GPT-4V's efficacy is close to ViT-B/32, illustrating its competitive yet varying performance across a spectrum of visual tasks.
It's noteworthy that, as per the GPT-4V documentation\footnote{https://platform.openai.com/docs/guides/vision}, the low-resolution version of the model scales images to 512$\times$512, while the high-resolution version scales to 2048$\times$2048. As Table~\ref{tab:datasets} illustrates, many of the datasets feature relatively lower resolutions, with the majority significantly below 512$\times$512. This discrepancy may impact GPT-4V's recognition accuracy, as seen in the case of the EuroSAT dataset, which has a resolution of 64$\times$64.

2) For video datasets, it’s important to highlight that Something-Something V1~\cite{sth-sth} focuses on modeling temporal relationships, whereas UCF101\cite{ucf101}, HMDB51\cite{hmdb}, and Kinetics~\cite{i3d} are less dependent on such temporal relationships, meaning actions can often be inferred from individual frames, as shown in Figure~\ref{fig:k400}. GPT-4V performs well on Kinetics, UCF101, and HMDB51, significantly surpassing EVA ViT-E's performance on UCF101 and HMDB51: achieving 83.7\% \vs 74.8\% on UCF, and an even more significant 63.2\% \vs 41.5\% on HMDB. This superior performance may be due to GPT-4V's adeptness at drawing inferences from the context of adjacent frames.
Notably, GPT-4V's performance on the SSV1 dataset is also markedly poor, at just 4.6\% top-1 accuracy, which aligns with the CLIP baseline. 
This is exemplified in Figure~\ref{fig:SSV1}, where isolating each frame does not provide enough context to ascertain the person's activity; only through the analysis of motion information across a sequence of frames can we make a prediction. Such results highlight the limitations of GPT-4V in temporal modeling due to the absence of a video encoder capable of processing temporal dynamics and motions.

3) For point cloud datasets, GPT-4V demonstrates excellent performance with just six rendered images, on par with EVA ViT-E. It stands to reason that adding more views would likely enhance the recognition accuracy even further.

In this section, all the above results are intended to provide baseline data and experience for future research, encouraging the development of more effective prompts to guide GPT-4V towards more accurate recognition.

\begin{figure}[t]
  \centering
   \includegraphics[width=1\linewidth]{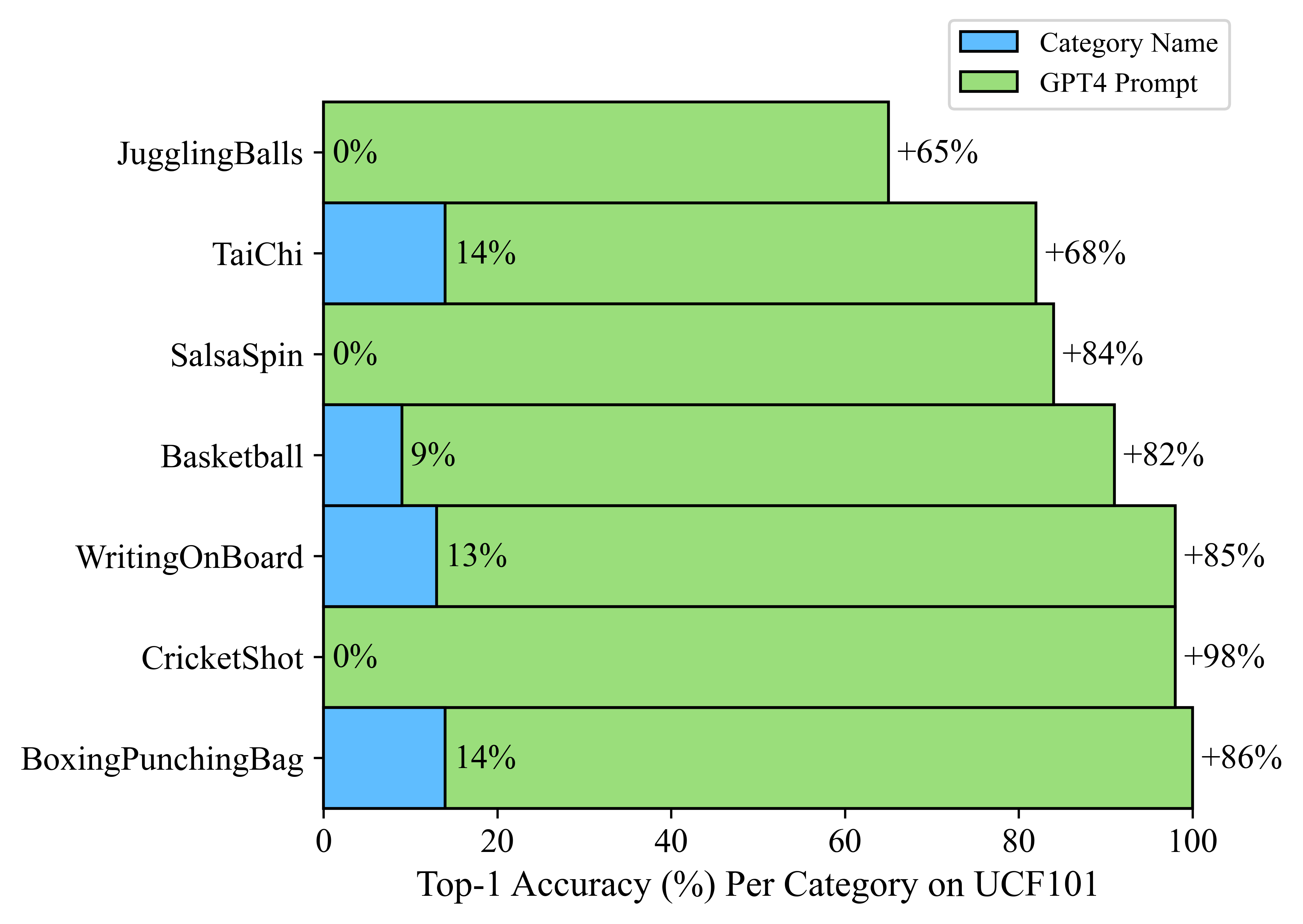}
   \vspace{-5mm}
   \caption{GPT Prompt \vs Category Name in some classes.} 
   \label{fig:clip_gpt4prompt}
\end{figure}

\subsection{Ablation Studies on \whblue{GPT Prompts}}
Here we present several ablation studies demonstrating the impact of prompts on CLIP's zero-shot performance.

\vspace{1mm}
\noindent\textbf{Impact of different prompts.}
Table~\ref{tab:clip_prompt} comprehensively exhibits the results of different prompts on the zero-shot visual recognition performance of CLIP across various datasets. Augmenting this with Hand-crafted Prompt combined with the category names leads to further improvements in most datasets, showcasing the method's robustness.
We then explore the effectiveness of employing multiple GPT-generated descriptive sentences related to category names. We find that GPT Prompt outperforms the baseline in 14 datasets. Figure~\ref{fig:clip_gpt4prompt} showcases the performance enhancement of GPT Prompts over category names in certain categories. 
Also, GPT Prompt can achieve better performance than Hand-crafted Prompts in 10 datasets. 
Our conjecture is that single category names may convey more global concepts, while the fine-grained details in generated descriptions are likely to align more closely with the visual content, thus amplifying inter-class distinctiveness. The strategy of generating multiple descriptive sentences may potentially further augment this effect. 

However, it's noteworthy that GPT Prompts are either below or roughly on par with Hand-crafted Prompt in 6 datasets, particularly in SUN397~\cite{xiao2010sun}, ImageNet-1k~\cite{deng2009imagenet}, Oxford Cars~\cite{Stanfordcar}, and Kinetics-400~\cite{i3d}. These datasets generally have a large number of categories with an emphasis on highly fine-grained classification. For such closely similar categories (like similar cars or scenes), richer descriptions generated may not be as distinctive as simply using the category name. Therefore, we consider combining ``Hand-crafted Prompt + GPT Prompts" to amalgamate the advantages of both, which has led to improved results in 11 datasets. For the 4 datasets (\ie, EuroSAT~\cite{helber2019eurosat}, RAF-DB~\cite{li2017reliable}, Flower102~\cite{flower} and ModelNet10~\cite{modelnet40}) where GPT Prompts demonstrate a clear advantage, the integration of Hand-crafted Prompt has been deemed unnecessary.

\begin{table}[t]
    \centering
    \scalebox{0.9}{
    \begin{tabular}{lcc}
    \toprule
     Method & \# Sentences   &  Top-1(\%) \\ \midrule
      Baseline: Category name   & 1 & 40.2  \\ \midrule
      \multirow{6}{*}{\whblue{\textbf{GPT Prompts}}}  & 1 & 38.4  \\
       & 3 & 42.9   \\
       & 5 & 47.4  \\
       & 10 & 49.1  \\
       & 20 & 49.4   \\
       & 30 & 49.9   \\
      \bottomrule
    \end{tabular}
    }
    \caption{The impact of different numbers of sentences generated by GPT on EuroSAT dataset. Backbone: CLIP ViT-B/32.}
    \label{tab:n_sen}
\end{table}

\vspace{1mm}
\noindent\textbf{Impact of sentence quantity generated by GPT.}
Our exploration also delved into the effect of the number of descriptive sentences generated by GPT-4 on zero-shot performance. 
Taking the EuroSAT~\cite{helber2019eurosat} dataset as an example, as shown in Table~\ref{tab:n_sen}, performance with only one generated sentence was lower than using the category name alone. However, an increase to three sentences led to a noticeable improvement and surpassed the baseline (42.9\% \vs 40.2\%). With five sentences, there was a substantial performance boost. In pursuit of identifying a saturation point for this improvement, we observed that increasing to 20 sentences brought about minimal additional benefits. Consequently, we adopt the generation of 20 sentences as the default setting for our experiments.

%% file: sec/4_limitation.tex
\section{Special Cases and Discussion on \textbf{\whred{GPT-4V}}}
In this section, we primarily present some special phenomena observed during the evaluation, provided for the reference of future researchers.

\begin{table}[!t]
    \centering
  \scalebox{0.82}{
  \setlength{\tabcolsep}{1.0pt}
    \begin{tabular}{lccccccccccc}
    \toprule
     & DTD  & SUN & RAF & Caltech & ImgNet & Aircraft & Flower & Cars & Food & Pets \\
    \midrule
    Batch     &  59.1  & 57.0 & 58.5 & 95.5  & 58.5 & 36.0 & 70.6 & 58.3 & 79.0  & 92.6 \\
    \rpink
    Single    &  57.7  & 59.2 & 68.7 & 93.7  & 63.1  & 56.6 & 69.1 & 62.7  & 86.2 & 90.8  \\
    \bottomrule
    \end{tabular}
    }
    \caption{Impact of Single \vs Batch Testing on GPT-4V performance evaluation, reporting top-1 accuracy (\%).}
    \label{tab:single}
\end{table}

\vspace{1mm}
\noindent\textbf{Batch Testing \vs Single Testing.} Our initial results were released before December 2023, during a period when we were constrained by OpenAI's daily API request limit of 100 per account. This led us to implement \emph{Batch Testing}, where one request yielded results for multiple samples.
Moving closer to March 2024, the limits were removed for tier 5 accounts, enabling us to switch back to standard \emph{Single Testing}, updating all datasets with one result per request. \Cref{tab:single} shows a comparison of these two methods. The results from batch testing do not align perfectly with those from standard single testing. This misalignment could stem from several factors: 1) There's a notable probability of result misalignment, \ie, the results intended for sample A may actually correspond to sample B. Furthermore, we've observed instances of both repeated predictions for the same sample and missing predictions for others.
2) Given that GPT-4V can process all samples in a batch simultaneously, the content of these samples might interfere with the results of individual samples.
Therefore, we recommend that future work should invariably employ single testing (\ie, the API processes only one sample at a time) to ensure the accuracy of the tests.

\vspace{1mm}
\noindent\textbf{Predict categories beyond the given list.}
In some instances, GPT-4V predicted categories that were not included in the given category list. For example, during the evaluation of the texture recognition dataset DTD~\cite{dtd}, GPT-4V might respond with: ``\emph{Note: As there were no categories provided that perfectly matched some images (such as straw-like), I have used the most comparable or related terms from the list provided. Additionally, not all images may have an exact or obvious matching category within the provided list, and I've estimated the most relevant categories based on observable texture patterns in the images.}'' In such cases, we tried to restrict the predictions to the provided category list through the prompts, but this proved ineffective. To proceed with the evaluation, we chose to exclude these predictions that were not within the given list.

\vspace{1mm}
\noindent\textbf{Prediction based on image names.}
GPT-4V sometimes makes category inferences from image names. For streamlined data gathering, we set GPT-4V to output in a dictionary format, with sample ID from filename as the key, and predictions as the value. 
Notably, GPT-4V may rely on explicit category hints in filenames, like `banded\_0060.jpg' from the DTD~\cite{dtd} dataset, impacting its predictions more than the visual content itself. 
or instance, the DTD dataset's top-1 accuracy soared to 98\% with original filenames but normalized to 57\% when filenames were anonymized.
To address this, we hashed sample names to ensure GPT-4V focuses on visuals over filename cues.

\begin{figure}[t]
    \centering
    \includegraphics[width=1\linewidth]{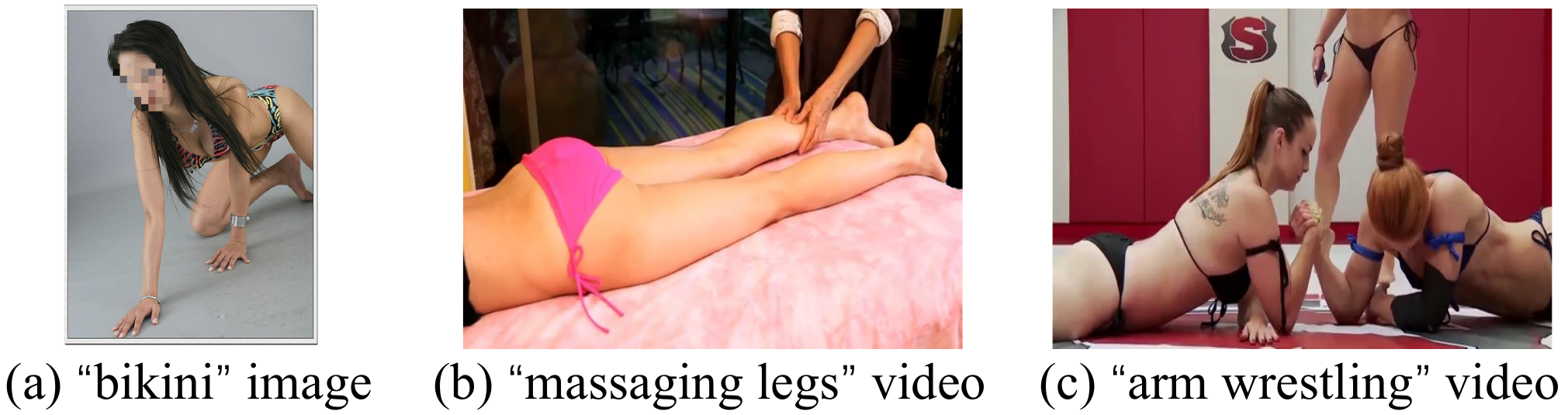}
    \caption{Some examples rejected by GPT-4V. (a) is an image from ImageNet-1K, (b) and (c) are videos from Kinetics-400.}
    \label{fig:safety}
\end{figure}

\vspace{1mm}
\noindent\textbf{Safety system in GPT-4V.}
Throughout our dataset evaluations, we stumbled upon specific instances, as depicted in Figure~\ref{fig:safety}, where GPT-4V refused to generate predictions, 
stating: ``\emph{Your input image may contain content that is not allowed by our safety system.}''
We surmise that this precautionary mechanism is designed to ensure that GPT-4V adheres to ethical guidelines by avoiding engagement with potentially sensitive or inappropriate content.

\vspace{1mm}
\noindent\textbf{GPT-4V API Costs.} We offer an estimate that using the GPT-4V API for one testing round across all datasets costs about \$4000 for reader's reference.

\section{Conclusion and Limitation}
This work aims to quantitatively evaluating the linguistic and visual capabilities of the current state-of-the-art large multimodal model GPT-4 in zero-shot visual recognition tasks. To ensure a comprehensive evaluation, we have conducted experiments across three modalities—images, videos, and point clouds—spanning a total of 16 benchmarks. We hope our empirical study and experience will benefit the community, fostering the evolution of future multimodal models.

\noindent\textbf{Limitations:} 1) This study has focused solely on fundamental visual recognition tasks. A comprehensive quantitative analysis of other tasks, such as object detection, is necessary to truly gauge the breadth of these models' capabilities in analyzing complex visual information. 
2) This work is limited to the evaluation of GPT-4 alone. Future efforts could include quantitative comparisons of various multimodal models (\eg, LLaVA~\cite{liu2023llava}, MiniGPT-4~\cite{zhu2023minigpt}, \etc), enhancing the breadth and depth of our analysis.
3) The current method of prompting GPT-4V is quite straightforward, raising concerns that an overly lengthy category list will increase the number of input tokens, potentially leading to adverse effects. Designing more effective prompts may further unlock GPT-4V's capabilities.
